\documentclass[lettersize,journal]{IEEEtran}
\usepackage{amsmath,amsfonts}
\usepackage{algorithmic}
\usepackage{algorithm}
\usepackage{array}
\usepackage[caption=false,font=normalsize,labelfont=sf,textfont=sf]{subfig}
\usepackage{textcomp}
\usepackage{stfloats}
\usepackage{url}
\usepackage{verbatim}
\usepackage{graphicx}
\usepackage{cite}
\graphicspath{{picture/}}
\usepackage{booktabs}
\usepackage{caption}
\usepackage{color}
\begin{document}

\title{Efficient Anti-exploration via VQVAE and Fuzzy Clustering in Offline Reinforcement Learning} 

\author{Long Chen,~\IEEEmembership{Senior Member,~IEEE,} Yinkui Liu, Shen Li, Bo Tang,~\IEEEmembership{Senior Member,~IEEE}, and Xuemin Hu 

\thanks{This work was supported by the National Natural Science Foundation of China (62273135, 62373356), the Joint Funds of the National Natural Science Foundation of China (U24B20162), the Natural Science Foundation of Hubei Province in China (2025AFA083), and Original Exploration Seed Project of Hubei University (202416403000001). (Corresponding author: Xuemin Hu.)}
\thanks{Long Chen is with the State Key Laboratory of Multimodal Artificial Intelligence Systems and the State Key Laboratory of Management and Control for Complex Systems, Chinese Academy of Sciences, Beijing, 100190, China. Long Chen is also with WAYTOUS Inc., Beijing, 100083, China, the Guangdong Laboratory of Artificial Intelligence and Digital Economy (SZ), Shenzhen 518107, China, and the Institute of Artificial Intelligence and Robotics, Xi'an Jiaotong University, Xi'an, 710049, China. (e-mail: long.chen@ia.ac.cn)}
\thanks{Yinkui Liu and Xuemin Hu are with the School of Artificial Intelligence, Hubei University, Wuhan 430062, Hubei, China, and also with the Key Laboratory of Intelligent Sensing System and Security (Hubei University), Ministry of Education, Wuhan, Hubei, 430062, China. (e-mail: huxuemin2012@hubu.edu.cn)}
\thanks{Shen Li is with the Shanghai Research Institute for Intelligent Autonomous Systems, Tongji University, Shanghai, 201210, China. (e-mail: lirunshen@tongji.edu.cn)}%
\thanks{Bo Tang is with the Department of Electrical and Computer Engineering, Worcester Polytechnic Institute, Worcester, MA, 01609, USA. (e-mail: btang1@wpi.edu)}

}


\markboth{This work has been submitted to the IEEE for possible publication. Copyright maybe transferred without notice.}
{Shell \MakeLowercase{\textit{et al.}}: A Sample Article Using IEEEtran.cls for IEEE Journals}


\maketitle

\begin{abstract}
Pseudo-count is an effective anti-exploration method in offline reinforcement learning (RL) by counting state-action pairs and imposing a large penalty on rare or unseen state-action pair data. Existing anti-exploration methods count continuous state-action pairs by discretizing these data, but often suffer from the issues of dimension disaster and information loss in the discretization process, leading to efficiency and performance reduction, and even failure of policy learning. In this paper, a novel anti-exploration method based on Vector Quantized Variational Autoencoder (VQVAE) and fuzzy clustering in offline RL is proposed. We first propose an efficient pseudo-count method based on the multi-codebook VQVAE to discretize state-action pairs, and design an offline RL anti-exploitation method based on the proposed pseudo-count method to handle the dimension disaster issue and improve the learning efficiency. In addition, a codebook update mechanism based on fuzzy C-means (FCM) clustering is developed to improve the use rate of vectors in codebooks, addressing the information loss issue in the discretization process. The proposed method is evaluated on the benchmark of Datasets for Deep Data-Driven Reinforcement Learning (D4RL), and experimental results show that the proposed method performs better and requires less computing cost in multiple complex tasks compared to state-of-the-art (SOTA) methods.
\end{abstract}


\section{Introduction}
\IEEEPARstart{A}{s} one of the most popular methods in modern cybernetics, offline reinforcement learning, which can learn polices from offline data collected by behavioral policies, avoids the cost and security issues caused by the interaction with the environment in traditional online RL methods when applied in high-risk exploration tasks such as autonomous driving \cite{chen2025Mixed, hu2023learning, hu2026Long}, and is increasingly drawing researchers' attention \cite{levine2020offline, Hu2026Enhancing,lian2021robust,liu2025ekg}. However, the offline datasets used for training often cannot cover the real data distribution in practical tasks, resulting in a distributional shift between the learned policy and the behavior policy \cite{lambert2022challenges}. In this case, the value function is prone to overestimate the out-of-distribution (OOD) data \cite{prudencio2023survey}, leading the agent to select the policies with high-risk or ineffective actions. In addition, unlike online RL, the agent in offline RL cannot correct the accumulative error caused by the overestimation due to the interaction with the environment, so it is a challenge to handle the overestimation issue of OOD data in offline RL\cite{huang2024offline}.

\begin{figure*}[!t]
    \centering
    \includegraphics[width=\textwidth]{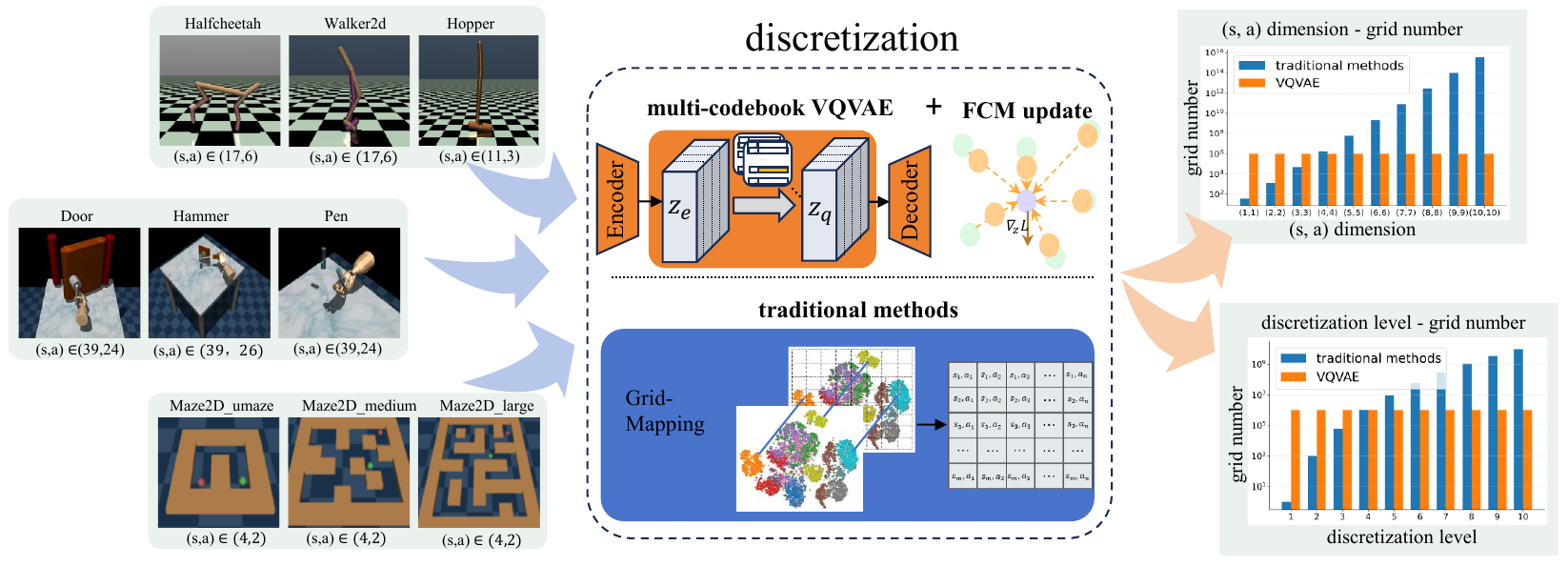}
    \caption{Comparison of the dimensions after discretization between the proposed method and traditional methods. Traditional pseudo-count methods discretize the continuous space into grids for counting, where the grid number greatly increases as the input state-action dimension and the discretization level, leading to dimension disaster. We propose a multi-book VQVAE-based pseudo-counting method to discretize continuous state-action pairs to discrete vectors as well as developing a FCM-based method to update the codebook vectors, effectively reducing the data dimension and information loss after discretization.}
    \label{VQVAEandFCM}
\end{figure*}

Anti-exploration is a conservative policy method which reduces the estimated Q-values or rewards for OOD data by imposing large penalties on unseen or rare data to address the overestimation issue and improve the stability of the learned policy. The key point is how to impose appropriate penalties. Calculating the penalty value based on the pseudo-count values of state-action pairs is a common and effective method for anti-exploration methods \cite{rezaeifar2022offline}, where pseudo-counting is used to quantify the uncertainty of state-action pairs by approximately evaluating the number of state-action pairs and imposing penalties on rare or unseen state-action pairs, thus alleviating the OOD problem. These anti-exploration methods based on counting state-action pairs can easily handle discrete data but fail to work in continuous state-action spaces since the number of state-action pairs could be infinite in continuous spaces. 

Recent studies have shown that counting state-action pairs after dividing the continuous state-action space into discretized grids is an intuitive and effective pseudo-count method \cite{shen2024grid}. However, these methods fail to effectively preserve distance information of features between similar states, and adjacent states may be divided into   different grids, generating information loss and inaccurate counting. Moreover, the grid number greatly increases with the dimensions of the state and action, which makes these anti-exploration methods for continuous state-action data suffer from the dimension disaster issue, leading to excessive computing cost, as shown in Fig. \ref{VQVAEandFCM}. 

To address these challenges, a novel anti-exploration method based on VQVAE \cite{van2017neural} and fuzzy clustering in offline RL is proposed in this paper. The proposed method transforms the counting for state-action pairs into the counting for the label sequences of the selected discrete vectors in codebooks during the VQVAE quantification process. Since these labels are present in the form of one-dimensional integers, the dimension of the label space no longer increases with the dimension of state-action pairs, thus effectively handling the dimension disaster issue. To enhance the expression ability of the VQVAE, we use the multi-codebook VQVAE and divide the encoded latent space into multiple subspaces. The feature vectors in these subspaces are separately quantized using the vectors in the corresponding codebooks. In addition, a new update method based on fuzzy C-means (FCM) clustering \cite{nascimento2000fuzzy, nguyen2015online} is also developed for vectors in codebooks. By calculating the membership degree between each vector in the divided potential subspaces and its corresponding vector in the codebooks, each vector in the divided potential subspaces can participate in the update process of multiple codebook vectors, thus improving the use rate of codebook vectors and reducing information loss in the discretization process. In conclusion, this paper offers the following three contributions.

\begin{itemize}
\item{We propose a pseudo-count method for discretizing state-action pairs based on vector labels in a multi-codebook VQVAE, where continuous state-action pairs are mapped to a discrete space and the state-action pairs are counted using the vector label sequences in the multi-codebooks quantitated by VQVAE, effectively reducing the computing cost.}


\item{A codebook update method based on fuzzy C-means clustering is developed to improve the use rate of vectors in codebooks and reduce information loss in the discretization process.}
\item{Comprehensively experiments were conducted on the D4RL benchmark for offline RL. The experimental results demonstrate that the proposed method outperforms SOTA offline RL methods in most tasks of D4RL.}
\end{itemize}

\section{Related Work}
\subsection{Anti-exploration Methods in Offline RL}
Rezaeifar et al. first introduce the concept of anti-exploration in offline RL \cite{rezaeifar2022offline}, which prevents the policy from excessively exploring OOD data by imposing a penalty on OOD data. After that, a few researchers followed this idea and conducted some studies in offline RL that are mainly divided into three categories: error-based penalty, behavioral policy constraints, and uncertainty-driven regularization.  

Error-based penalty methods for anti-exploration have achieved great success in recent years. TD3-CVAE \cite{rezaeifar2022offline} and SAC-RND \cite{nikulin2023anti} use the reconstruction error of Conditional Variational Autoencoder (CVAE) and the prediction error of Random Network Distillation (RND), respectively, as the penalty signal to effectively handle the extrapolation error issue caused by OOD data. Zhang et al. prove that the ``anti-exploration rewards" contributes to policy consistency \cite{zhang2021made}. The methods based on behavioral policy constraints limit the deviation of the learned policy from the behavior policy through policy constraints or budget mechanisms \cite{fujimoto2019off, fujimoto2021minimalist, kumar2019stabilizing, mao2024odice}, which improve policy safety and stability without sacrificing sample efficiency. As a representative method, O-DICE \cite{mao2024odice} projects the backward gradient onto the normal plane of the forward gradient to strengthen constraints into state-action data, and achieves good results in the field of continuous control. The methods based on uncertainty-driven regularization measure the intensity of OOD data and impose penalties on the actions with high uncertainty, leading to accurate identification for OOD data \cite{an2021uncertainty, shen2024grid, nikulin2023anti, rezaeifar2022offline}. As a classic method, Conservative Q-Learning (CQL) \cite{arasu2006cql} directly suppresses the estimated Q-values of OOD actions in the value function, to deal with the overestimation issue of Q-values.

Following the idea of the third method, our method imposes a regularization constraint on the policy by calculating the uncertainty of OOD data as a penalty. Compared to existing methods, we use a more intuitive and simple way, where we propose a novel pseudo-count method based on multi-codebook VQVAE to count state-action pairs and use the counting result as the penalty value.

\subsection{Pseudo-count Methods}
Pseudo-counting denotes that some features in high-dimensional or continuous state spaces that cannot be directly counted are approximately estimated by the access number of other features \cite{li2013accurate}. In online RL, it is often used to measure the novelty of states or state-action pairs, thus calculating the exploration rewards \cite{bellemare2016unifying, tang2017exploration, machado2020count}. Bellemare et al. use PixelCNN to estimate the probability distribution of observed data and indirectly obtain pseudo-count values through its variations, encouraging agents to explore new states with low pseudo-counts, which achieves good results in Atari games \cite{ostrovski2017count}. 

In recent years, some researchers introduced pseudo-count into offline RL to suppress excessive utilization of OOD state-action pairs by polices. Kim et al. map continuous or high-dimensional states into binary vectors through hash encoding to achieve approximate counting and impose conservative penalties on low-counted state-action pairs, but its accuracy was limited by the hash function \cite{kim2023model}. To improve counting accuracy, Shen et al. propose GPC-SAC which discretize the continuous space into grids for indirect counting \cite{shen2024grid}. However, the grid number is extremely large, making it difficult to count in high-dimensional states, and the uniform discretization in this method easily allocates adjacent data to different grids, resulting in information loss and counting errors.

The method most similar to ours is GPC-SAC. The main difference is that we convert counting in the latent space to counting the label sequences of selected codebook vectors in multi-codebook of VQVAE rather than counting in the discretized grids during quantization, to effectively reduce computing cost and information loss, thus improving the accuracy of counting state-action pairs for stable policy learning.

\section{Preliminaries}
\subsection{Anti-Exploration Based on Pseudo-count}
The objective of offline RL is to optimize the policy $\pi_\theta$ based on the offline dataset $D = \{(s_t, a_t, r_t, s_t')\}$ that is generated by the behavior policy $\pi_\beta$ and contains the trajectories consisting of the state \(s\), action \(a\), the corresponding reward \(r\), and the transferred state \(s'\) without interacting with the environment. The subscript $t$ denotes the time step. In offline RL methods, pseudo-counting is often used to correct incorrect Q-values or rewards by calculating a penalty value that is usually obtained by approximating the occurrence frequency of state-action pairs using probability density models or discrete counting \cite{bellemare2016unifying, tang2017study}. Assuming that $\beta > 0$ is the penalty coefficient, the correction of the Q-value based on the pseudo-count value can be expressed by Eq. \ref{eq:1}.
\begin{equation}
Q_{ood}(s,a)=Q(s,a)-\frac{\beta}{\sqrt{n(s,a)}},
\label{eq:1}
\end{equation}
where $Q(s,a)$ and $Q_{ood}(s,a)$ denote the Q-values of the state–action pair $(s,a)$ before and after subtracting the penalty term, respectively. $n(s,a)$ denotes the pseudo-count value, and a low $n(s,a)$ denotes a small penalty and contributes to constrain the policy to excessively explore the distribution regions of sparse data, thus reducing the overestimated Q-values for OOD data. $\beta$ is a hyperparameter that adjusts the penalty intensity. As the pseudo-count number increases in the training process, the penalty value decreases, which allows the policy to progressively explore rare state–action pairs while keeping stable policy learning performance, thus achieving good generalization for the offline data as well as OOD data.

\subsection{VQVAE}
Vector Quantized Variational Autoencoder, an encoding method that discretizes continuous latent spaces by introducing a vector quantization mechanism, can learn an effective representation for discrete features while preserving the representation ability of continuous encoders. It consists of three main parts: an encoder $E$, a decoder $D$, and a codebook $\{e_k\}_{k=1}^N$, where $N$ is the element number in the codebook and each element $e_k$ is represented by a discrete latent vector. The encoder first maps the input data $x$ to a continuous latent representation. The distances between these continuous vectors mapped from the input data and the discrete codebook elements $e_k$ are then calculated as Eq. \ref{eq:2}, and each continuous vector $z_e$ is replaced by the closest codebook vector $e_k$ to itself, which achieves a discrete representation of the continuous latent space.
\begin{equation}
z_q^k = \arg \min_{1 \leq k \leq N} \left\| z_e^k - e_k \right\|_2.
\label{eq:2}
\end{equation}

To jointly train the encoder, codebook, and decoder, the overall loss function $\mathcal{L}_{VQ}$ of VQVAE consists of three terms, as defined by Eq. \ref{eq:3}.
\begin{equation}
\begin{split}
    \mathcal{L}_{VQ} = & \left\| x - D(z_q) \right\|^2 
    + \left\| \text{sg}[E(x)] - e_k \right\|^2 \\
    & + \gamma \left\| E(x) - \text{sg}[e_k] \right\|^2,
\end{split}
\label{eq:3}
\end{equation}

where $x$ and $z_q$ denote the input data and the discretized latent vector by VQVAE, respectively. $sg[.]$ denotes the stop-gradient operation, which means that the term is excluded from gradient update during backpropagation. In Eq. \ref{eq:3}, the first term represents the reconstruction loss, which is used to ensure that the decoder can reconstruct the input $x$ from the discretized latent vector $z_q$. The second term is the codebook loss, which makes the discrete codebook vector $e_k$ close to the encoder output $z_e$, keeping the consistency between the codebook and the continuous latent space. The third term is the commitment loss, which makes the encoder output $z_e$ close to its corresponding codebook vector, preventing $z_e$ from differing too much from the codebook vector after quantization. 

\subsection{Counting Bloom Filter}
Counting Bloom Filter (CBF) combines a Bloom filter with counters to keep track of the occurrence numbers of data, and is a data processing tool with high space efficiency and fast query time \cite{li2013accurate}. CBF often contains $N_c$ counters $C[1], C[2], ..., C[N_c]$, and its storage memory is smaller than that of hash tables that store complete counting objectives. The parallel processing ability of CBF can also reduce the cost of the counting time. Every time the state-action pair $(s,a)$ occurs, CBF assigns $e$ independent counters to this input datum, and each counter uses an independent hash function to calculate the corresponding counter index $i$. Then, the values of all the $e$ counters are added by 1, as shown by Eq. \ref{eq:CBF}. 
\begin{equation}
C(i) \leftarrow C(i) + 1, \quad i\in\{1,2,...,N_c\}.
\label{eq:CBF}
\end{equation}

When counting the occurrence number of the state-action pair $(s,a)$, the counting result, which is also the pseudo-count value $n(s,a)$, is equal to the minimum value of all counters indicated by the hash values, as shown by Eq. \ref{eq:min}.
\begin{equation}
n(s, a) = \min_{i \in \{1,2,...,N_c\} } C(i).
\label{eq:min}
\end{equation}

\section{Methodology}
In the proposed anti-exploration method for offline RL, we first propose a novel multi-codebook VQVAE-based pseudo-count method for discretizing state-action pairs, which maps input continuous state-action pairs into vectors in a discrete space. And then, we develop an anti-exploration method based on the proposed pseudo-count method combing the SAC algorithm \cite{haarnoja2018soft}, where the counting values serve as the dynamic penalty term to limit the SAC algorithm from overestimating the Q-values for rare or sparse state-action pairs. In addition, a discrete codebook vector update method based on FCM clustering is developed to improve the use rate of codebook vectors and reduce information loss in the quantization process, enhancing the performance of RL policy. 

\subsection{Pseudo-counting Based on Multi-codebook VQVAE}

\begin{figure*}[tb!]
	\centering
	\includegraphics[width=1\textwidth]{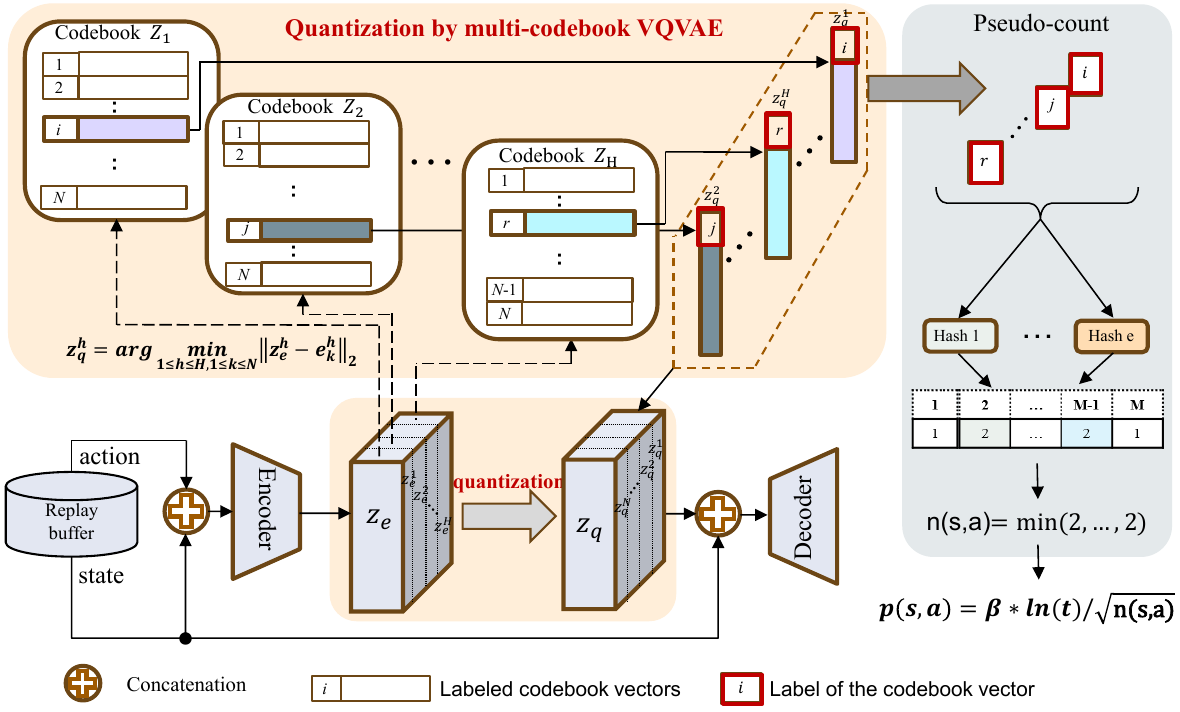} 
	\caption{Pseudo-counting based on multi-codebook VQVAE During quantization, a vector is selected from each codebook, and its corresponding vector label is obtained, forming a one-dimensional sequence of integer labels. Subsequently, a pseudo-counting is performed on this label sequence to calculate the corresponding penalty value $p(s, a)$.}
	\label{multi_codebook}
\end{figure*}

Due to the limited accuracy of counting functions, existing pseudo-count methods are difficult to distinguish the differences in various compressed features in the encoded latent space, leading to miscounts. VQVAEs use a codebook with fixed size for quantization, where the input data are mapped by selecting the nearest codebook vectors to themselves\cite{van2017neural}. However, for the input data represented by one-dimensional state-action pairs, the encoder in VQVAE outputs one-dimensional vectors, and the quantization is performed on a single vector \cite{ozair2021vector}. According to the quantization mechanism in the VQVAE, only one vector will be selected from the codebook to discretize input data, which means that each vector in the codebook needs to represent a large amount of information to make the codebook cover the state-action distribution when the traditional VQVAE with only a codebook is applied to discretize state-action pairs in complex continuous control tasks, resulting in insufficiently fine representation for each vector and information loss during quantization. Increasing the vector number in the codebook can alleviate this problem but significantly increase the computing cost.

Previous studies demonstrate that, under the same computing cost condition, increasing the codebook number can more effectively improve the expression of the encoder than increasing the vector number in a codebook in VQVAEs \cite{takida2023hq, luu2024mitigating}. Inspired by this idea, we propose a novel pseudo-count method based on a multi-codebook VQVAE, where multiple independent codebooks in the VQVAE are used to discretize high-dimensional state-action pairs, and each codebook focuses on learning fine-grained expression. Compared to a single codebook, the encoder with multiple codebooks for multidimensional and parallel discretization can represent the distribution of state-action pairs in high-dimensional spaces with finer granularity at the same computing cost, thus reducing quantization errors and improving accuracy of pseudo-counting. In this case, the latent space $z_e$ encoded by the VQVAE is uniformly sequentially partitioned into $H$ contiguous subspaces, and each subspace contains $N/H$ latent vectors and is independently quantized using its special codebook, as shown by Eq. \ref{eq:4}.


\begin{equation}
z_e = \left[ z_e^1, z_e^2, \ldots, z_e^H \right],
\label{eq:4}
\end{equation}

Each subspace $z_e^h \quad (h=1,2,...,H)$ is individually quantized with the corresponding codebook, so $H$ is also the codebook number in the multi-codebook VQVAE. In order to map continuous features to discrete vectors, similarly to Eq.\ref{eq:2}, the codebook vector with the closest distance in the corresponding codebook is selected for quantization, as shown by Eq. \ref{eq:5}.

\begin{equation}
z_q^h = \arg \min_{1 \leq h \leq H, 1 \leq k \leq N}\left\| z_e^h - e_k^h \right\|_2,
\label{eq:5}
\end{equation}
where $e_k^h$ denotes the \( k \)-th vector in the \( h \)-th codebook. $N$ is the total number of vectors in a codebook. For the latent space vector $z_e^h$, $z_q^h$ is the selected vector in the \( h \)-th codebook. The selected vectors from each codebook are then concatenated in sequence and form the quantization result $z_q$ that contains $H$ vectors, as shown by Eq. \ref{eq:6}.
\begin{equation}
z_q = \left[z_q^1, z_q^2, \ldots, z_q^H \right].
\label{eq:6}
\end{equation}

To effectively represent the dependencies of state-action pairs, the state is used as an additional condition input in our method following the structure of Conditional VAE \cite{zhou2021plas}. As shown in Fig. \ref{multi_codebook}, the state is first concatenated with the action and jointly fed into the encoder. The latent representation outputted from the encoder is subsequently quantized by the multi-codebook VQVAE to obtain the latent feature $z_q$. During the decoding stage, $z_q$ is again concatenated with the state, serving as the input to the decoder.

Similar to Eq. \ref{eq:3}, the optimization objective for the multi-codebook VQVAE in our method is also composed of three terms: the reconstruction error loss, the codebook update loss, and the commitment loss. The difference in the objective between our method and the traditional VQVAE is that the codebook update loss and the commitment loss in our method are the sums of the codebook update losses and the commitment losses of all codebooks, respectively, to ensure the update for all codebooks, as shown in Eq. \ref{eq:7}.

\begin{align}
\mathcal{L} = & \; \| a - \hat{a} \|^2 
+ \sum_{h=1}^{H} \| \operatorname{sg}\!\left[ z_e^h \right] - e_k^h \|^2 \nonumber \\
& + \gamma \sum_{h=1}^{H} \| z_e^h - \operatorname{sg}\!\left[ e_k^h \right] \|^2,
\label{eq:7}
\end{align}
where $\gamma$ is the coefficient for the commitment loss that is set to 0.25 following \cite{van2017neural}, and is used to balance the optimization between the encoder and the codebook.

After discretizing input state-action pairs and obtaining corresponding discrete codebook vectors, counting state-action pairs is replaced by counting the discrete codebook vectors in our method. However, directly counting the discrete codebook vectors consumes a lot of computing resources since these vectors are usually floating-point vectors with high dimensions. Furthermore, existing counters are difficult to meet the accuracy requirement of distinguishing between similar vectors.

To handle this issue, we design an efficient counting method based on the label sequences of codebook vectors. As shown in Fig. \ref{multi_codebook}, there are a total of $H$ codebooks, each of which contains $N$ vectors labeled from 1 to $N$. 
As shown by the red rectangles in Fig. \ref{multi_codebook}, the $H$ labels of $z_q^1, z_q^2, ..., z_q^H$ from the $H$ codebooks form a one-dimensional label sequence, which serves as the discretized representation of the corresponding state-action pair. Compared to codebook vectors, label sequences are lower-dimensional and consist of integers, which can reduce both computational and storage costs. Moreover, counting the discrete labels can enhance the distinguishability between similar codebook vectors, thus improving the accuracy of pseudo-counting. 

After obtaining the label sequence, the CBF method \cite{li2013accurate} is then used to perform pseudo-counting and output the pseudo-count value $n(s,a)$ for the state-action pair $(s,a)$, as shown in Fig. \ref{multi_codebook}.

\subsection{Anti-exploration Based on Pseudo-Count}
In the above subsection, the count value $n(s, a)$ for a given state-action pair is obtained by CBF and subsequently used to calculate the penalty value. Unlike the penalty value as shown by Eq. \ref{eq:1}, a new penalty value $p(s,a)=\beta ln(t) / {\sqrt{n(s,a)}}$ is designed, where the training time step $t$ is taken into account in logarithms, since the penalty value needs to be larger in the early stage to learn conservative policies and smaller in the later stage to avoid excessive conservatism. Therefore, the Q-value correction method for anti-exploration in Eq. \ref{eq:1} is changed by Eq. \ref{eq:11} in our method.

\begin{equation}
Q_{ood}(s,a)=Q(s,a)-\frac{\beta ln(t)}{\sqrt{n(s,a)}},
\label{eq:11}
\end{equation}
where the value of $\beta$ is set to different values in different environments following our baseline method GPC-SAC \cite{shen2024grid}. Through Eq. \ref{eq:11}, a low-frequency occurring state-action pair receives a higher penalty value $p(s,a)$. As the frequency of a state-action pair increases, the penalty value $p(s,a)$ gradually decreases, allowing the agent to explore the regions of sparse state-action pairs and balance exploration and conservatism in policy learning, thus obtaining a robust policy. For this purpose, the OOD loss in the critic network is designed by Eqs. \ref{eq:12} and \ref{eq:13}.

\begin{equation}
q^f_{\text{ood}} = \left[ Q_{\theta_i}(s, a_{\text{new}}),\ \overline{Q}_{\theta_i}(s', a'_{\text{new}}) \right],
\label{eq:12}
\end{equation}
\begin{equation}
q^f_{\text{target}} = 
\left[
    \max\!\left(Q_{\theta_i} - p,\ 0\right),\;
    \max\!\left(\overline{Q}_{\theta_i} - 0.1p',\ 0\right)
\right],
\label{eq:13}
\end{equation}
where $a_{new}$ and $a_{new}^{'}$ denote the output actions from the policy network at the current and next time steps, respectively. $p'(s', a'_{\text{new}})$ is the penalty value for the state-action pair at the next time step, performing a conservative estimation for Q-values of rare or unseen state-action pairs. $q_{ood}^f$ and $q_{target}^f$ denote the Q-value of the action sampled by the current policy and the predicted value of the target Q-network at the next time step, respectively. $p$ and $p'$ are the abbreviations of $p(s,a)$ and $p'(s', a'_{\text{new}})$, respectively. $Q_{\theta_i}$ and $\bar{Q}_i $ denotes the predicted values of the \(i\)-th Q-network and \(i\)-th target Q-network, respectively, where $i=\{1,2\}$ following \cite{shen2024grid}.    

The final objective of the proposed anti-exploration method for offline RL is shown by Eq. \ref{eq:14}.
\begin{equation}
\mathcal{L} =
\mathbb{E}\!\left[ \left( q^f_{\text{ood}} - q^f_{\text{target}} \right)^2 \right]
+ \mathbb{E}\!\left[ \left( Q_{\theta_i}(s, a) - y \right)^2 \right].
\label{eq:14}
\end{equation}
The Q-value, adjusted by the penalty derived from the pseudo-couning, is denoted as $q^f_{\text{target}}$. The mean squared error (MSE) is used to calculate the loss. \(y\) represents the standard Q-target value. The policy network is updated by maximizing the expected return with entropy regularization. The parameters in all Q-networks and their corresponding target networks are updated via soft update for stability and convergence during training, which maintains fitting accuracy on the input data while preserving conservatism for unseen data. The specific process is detailed in Algorithm 1.

\begin{algorithm}[H]
\caption{Anti-exploration based on the multi-codebook VQVAE for offline RL}
\label{alg:vqvae_anti_exploration}
\textbf{Initialize:} Pretrained the multi-codebook VQVAE, policy network \(\pi_\phi\), Q-networks \(Q_{\theta_1},Q_{\theta_2}\), target Q-networks \( \bar{Q}_{\theta_1}, \bar{Q}_{\theta_2}\), pseudo-count container \(C\), temperature parameter \(\alpha \)
\begin{algorithmic}[1]
\STATE \textbf{for} epoch = 1 to Max(Epoch) \textbf{do}
    \STATE \hspace{0.5cm} Sample minibatch \( B = \{s, a, r, s'\} \) from dataset
    \STATE \hspace{0.5cm} compute $\log \pi$, adjust $\alpha$
    \STATE \hspace{0.5cm} Sample actions \( a_{\text{new}}, a'_{\text{new}} \sim \pi_\phi \)
    \STATE \hspace{0.5cm} Pseudo counting: \( n(s, a),n'(s', a'_{\text{new}})\)
    \STATE \hspace{0.5cm} Estimate \( p(s, a), p'(s', a'_{\text{new}}) \) by Eq. \ref{eq:11}
    \STATE \hspace{0.5cm} Update Q-networks:
    \[
        Q_{\theta_i} \gets Q_{\theta_i} - \eta_Q \nabla L_{\pi}
    \]
    \STATE \hspace{0.5cm} Update the policy network with loss:
    \[
        L_\pi = \mathbb{E}[\alpha \log \pi - Q], \quad \phi \gets \phi - \eta_\pi \nabla L_\pi
    \]
    \STATE \hspace{0.5cm} Soft update of target Q-networks:
    \[
        \bar{\theta}_i \gets \tau \theta_i + (1 - \tau) \bar{\theta}_i
    \]
\STATE \textbf{end for}
\end{algorithmic}
\end{algorithm}

\subsection{Codebook Update Based on FCM Clustering}
Conventional VQVAEs map continuous latent space vectors to discrete codebook vectors through nearest-neighbor lookup, which blocks direct gradient backpropagation to the codebook. Although the Straight-Through Estimator (STE) can be used to approximate gradient backpropagation to the encoder network, only the currently selected codebook vectors are updated, ignoring unselected vectors without gradient feedback. In this case, some vectors in the codebook are rarely updated, which significantly decreases the use rate of codebook vectors, making the codebooks difficult to cover the distribution of state-action pairs.
Low-use rate codebooks cannot provide sufficient vectors to represent and distinguish features with minor differences in the latent space, which increases quantization errors and information loss of the encoded features.


In fact, codebook vectors can be regarded as cluster centers in the latent space. 
Vector update in conventional VQVAEs, which relies on single-point mapping, leads to information loss and uneven updates among cluster centers \cite{zheng2023online}. In contrast, FCM allows each data point to be associated with multiple cluster centers with different membership degrees, enabling all cluster centers to update to different extents during each clustering iteration\cite{yang2022linear}. 
Inspired by this idea, we propose a codebook update method combining Fuzzy C-Means clustering with the conventional update mechanism of VQVAEs, which leverages the membership degrees of latent vectors to update the codebook vectors. The proposed FCM-based codebook update method allows each vector in the latent space to influence all codebook vectors to varying degrees, thereby improving the use rate of codebooks. 
\begin{figure*}[tb!]
	\centering
	\includegraphics[width=1\textwidth]{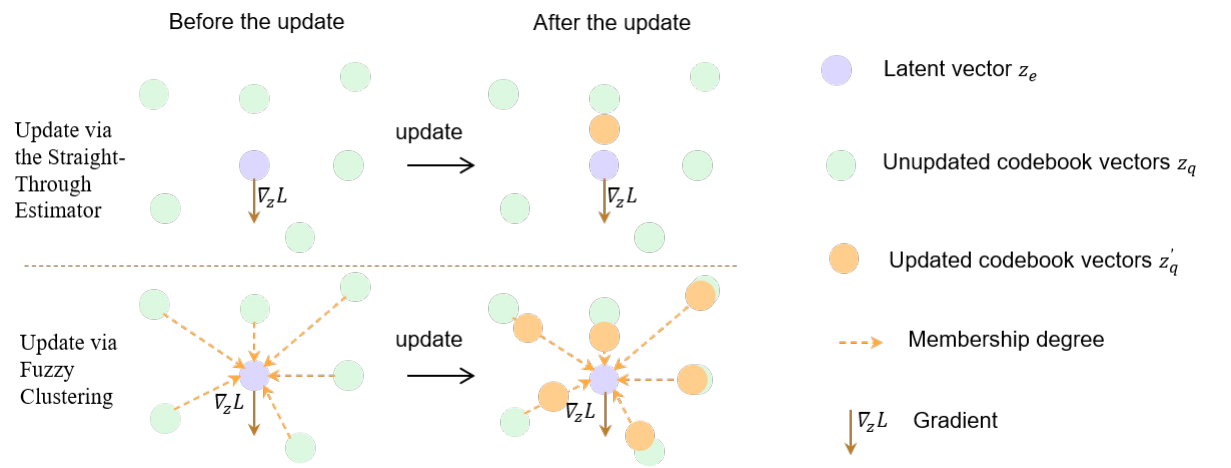} 
	\caption{Update process of FCM clustering. The membership degrees are calculated based on the Euclidean distances between the latent vectors and all the codebook vectors, and serve as the weights of corresponding codebook vectors in the update process, enabling joint optimization of all codebook vectors.}
	\label{Fuzzy_clustering_update}
\end{figure*}

The update process of FCM clustering is shown in Fig. \ref{Fuzzy_clustering_update}. In the codebook update process, the membership degree $f_{hk}$, which represents the degree of the feature vector $z_e$ belonging to the $k$-th vector $e_k$ in the $h$-th codebook , is calculated for each feature vector $z_e$ and each codebook vector $e_k$ in the corresponding codebook, as shown by Eq. \ref{eq:15}.

\begin{equation}
f_{hk} = \frac{d_k^{-2}}{\sum_{k=1}^{N} d_k^{-2}}.
\label{eq:15}
\end{equation}


Following the settings for the membership degree in \cite{zhao2021general, yu2015analysis, borgelt2013objective}, $N$ denotes the total vector number in each codebook. $d_k$ denotes the Euclidean distance between the feature vector $z_e$ and the $k$-th codebook vector $e_k$. Details of the parameters are provided in Table I in Section V.

To increase the update magnitude of infrequently used vectors, the update magnitude for each vector is calculated based on the use rate \( R_k \) of the \(k\)-th vector in a codebook, as shown by Eq. \ref{eq:16}.

\begin{equation}
\alpha = \exp\!\left(-\frac{10 N R_k}{1 - \varepsilon} - 10^{-3}\right),
\quad 1 \leq k \leq N,
\label{eq:16}
\end{equation}

where \( \epsilon \) is a decay hyperparameter that is set to 0.99 following \cite{zheng2023online}. $R_k=N^u_k/N^a_{k}$, where $N^u_k$ denotes the number of times that the \(k\)-th vector has been used. $N^a_{k}$ denotes the total number of times that all the vectors in a codebook have been used during the update process. To encourage the codebook vectors to cover the data distribution and increase the selected probability of infrequently used vectors, the codebook vectors are updated by the weighted means of normalized membership degrees. By multiplying the quantized feature vector $z_e$ with its corresponding membership degree $f_{hk}$, the \(k\)-th vector $e_k^h$ in the \(h\)-th codebook is iteratively updated based on the step size $\alpha$. This process is shown by Eq. \ref{eq:17}.
\begin{equation}
e_k^h \leftarrow e_k^h \cdot (1 - \alpha) + f_{hk} \, z_e \cdot a, \quad 1 \leq  k \leq N.
\label{eq:17}
\end{equation}

\section{Experiments}
\subsection{Experimental Settings}
To evaluate the proposed method, we select the Datasets for Deep Data-Driven Reinforcement Learning (D4RL) as the benchmark, which are collected from various kinds of environments and simulators including Maze2D, AntMaze, Adroit, FrankaKitchen, Flow, CARLA, as well as several Gym-MuJoCo tasks such as HalfCheetah, Hopper, and Walker2D. The used benchmark covers a variety of typical scenarios for offline RL testing, such as navigation, control, driving, and robot movement. The data in this benchmark are collected by different levels of policies, including expert, intermediate, and random, which cover significantly different distributions and are very suitable to evaluate the methods of handling the OOD issue, and some SOTA offline RL methods such as SAC \cite{haarnoja2018soft}, TD3+CVAE \cite{rezaeifar2022offline}, CQL \cite{arasu2006cql}, BCQ \cite{fujimoto2019off}, and IQL \cite{kostrikov2021offline} all use D4RL as the benchmark.

All experiments except for the hopper and walker2d tasks in the Gym environment were performed on a NVIDIA GeForce GTX 1080Ti. To accelerate training, an additional NVIDIA GeForce RTX 4090 was used to conduct the experiments in the hopper and walker2d tasks, with a total of 3000 for training epochs. Experimental evaluations were performed on five random seeds. Detailed settings are shown in Table \ref{tb:1}.

\begin{table}[!t]
  \centering
  \caption{Parameter settings used in experiments}
  \resizebox{\linewidth}{!}{
  \begin{tabular}{lrl}
    \toprule
    \textbf{Parameter} & \textbf{Value} & \textbf{Description} \\
    \midrule
    $d_{lat}$   & 64   & Latent space dimensionality \\
    $N$         & 256  & Number of vectors in a codebook \\
    $H$         & 4    & Number of codebooks \\
    $\gamma$    & 0.25 & Commitment loss coefficient \\
    $\text{lr}$ & 0.001& Initial learning rate \\
    \bottomrule
  \end{tabular}
  }
\label{tb:1}
\end{table}

\subsection{Metrics and Baselines}
We use the normalized score in D4RL as the main evaluation metric following some SOTA methods such as TD3-CVAE \cite{rezaeifar2022offline} and CQL \cite{arasu2006cql}. The average scores and variances of 10 consecutive tests are taken as the final metric. 

To evaluate the performance of our method, the proposed method is compared with three SOTA anti-exploration methods including TD3-CVAE \cite{rezaeifar2022offline}, SAC-RND \cite{nikulin2023anti}, and GPC-SAC \cite{shen2024grid}, as well as four mainstream offline RL methods including DMG \cite{mao2024doubly}, O-DICE \cite{mao2024odice}, IQL \cite{kostrikov2021offline}, and CQL \cite{arasu2006cql}.

\subsection{Experimental Results Compared to SOTA Methods}
Table \ref{tb:2} shows the scores of the comparative methods in different Gym datasets. Our method achieves the highest average score in 9 Gym tasks, and outperforms the benchmark method GPC-SAC in 14 tasks, where the average score of the proposed method is approximately 6.1\% higher than GPC-SAC, because the proposed method can more effectively represent the relevant information of the input states and perform a more accurate pseudo-count estimate, effectively mitigating overestimation and improving the policy accuracy.
The proposed method is significantly superior compared to TD3-CVAE and achieves comparable performance compared to SAC-RND, where TD3-CVAE and SAC-RND use the penalty to constrain the OOD data and avoid overestimation. Compared with the latest method DMG which restricts overvalues by reducing the propagation of generalization errors, the proposed method also achieves higher scores in multiple Gym tasks, where the average scores are about 18\% higher in some halfcheetah datasets such as halfcheetah-random, halfcheetah-medium, and halfcheetah-medium-replay. The proposed method exhibits relatively large score variances in some tasks with unevenly distributed data such as hopper-medium-v2 and hopper-medium-expert-v2 although obtaining high scores, because our method is overly conservative in penalizing the OOD data, leading to performance fluctuation.

\begin{table*}[htbp]
  \centering
  \caption{Evaluation on the Gym Datasets. ``ho" denotes Hopper. ``half" denotes HalfCheetah. ``w" denotes Walker2d. ``r," ``m," ``m-r," and ``e" represent random, medium, medium-replay, and expert, respectively. The scores of SAC-CVAE, IQL, CQL, and GPC-SAC are from \cite{shen2024grid}, while the scores for SAC-RND, DMG, and O-DICE are from \cite{nikulin2023anti}\cite{mao2024doubly}\cite{mao2024odice}.}
  \renewcommand{\arraystretch}{1.2}
  \setlength{\tabcolsep}{12pt}
  \begin{tabular}{lcccccccc}
    \toprule
    \textbf{Tasks} & \textbf{TD3-CVAE} & \textbf{SAC-RND} & \textbf{DMG} & \textbf{O-DICE} & \textbf{IQL} & \textbf{CQL} & \textbf{GPC-SAC} & \textbf{OURS} \\
    \midrule
    ho-r     & 11.7±0.2 & 31.3±0.1 & 20.4±10.4 & —         & 7.8±0.4 & 16.4±14.5 & 31.4±0.0 & \textbf{32.2±0.4} \\
    ho-m     & 55.9±11.4 & 97.8±2.3 & \textbf{100.6±1.9} & 86.1±4.0 & 66.2±5.7 & 53.0±28.5 & 82.9±2.3 & 97.1±12.1 \\
    ho-m-r   & 46.7±17.9 & 100.5±1.0 & 101.9±1.4 & 99.9±2.7 & 94.7±8.6 & 35.9±3.7 & 97.5±3.6 & \textbf{103.1±0.8} \\
    ho-m-e   & \textbf{111.6±2.3} & 109.8±0.6 & 110.4±3.4 & 110.8±0.2 & 91.5±14.3 & 105.6±12.9 & 109.8±3.4 & 102.9±19.0 \\
    ho-e     & —         & 109.7±0.3 & 111.5±2.2 & —         & 109.4±0.5 & 96.5±28.0 & 109.5±4.0 & \textbf{112.8±2.5} \\
    half-r   & 28.6±2.0 & 29.0±1.5 & 28.8±1.3 & —         & 11.0±2.5 & 28.3±0.5 & 31.0±1.8 & \textbf{33.8±0.9} \\
    half-m   & 43.2±0.4 & 66.6±1.6 & 54.9±0.2 & 47.4±0.2  & 47.4±0.2 & 47.0±0.5 & 60.8±0.7 & \textbf{68.2±0.8} \\
    half-m-r & 45.3±0.4 & 54.9±0.6 & 51.4±0.3 & 44.0±0.3  & 44.2±1.2 & 45.5±0.7 & 55.7±1.0 & \textbf{62.0±1.1} \\
    half-m-e & 86.1±9.7 & \textbf{107.6±2.8} & 91.1±4.2 & 93.2±0.6  & 86.7±5.3 & 75.6±25.7 & 87.1±5.8 & 104.9±2.3 \\
    half-e   & —         & 105.8±1.9 & 95.9±0.3 & —         & 95.0±0.5 & 96.3±1.3 & 104.7±2.2 & \textbf{108.8±0.8} \\
    w-r      & 5.5±8.0  & \textbf{21.5±0.1} & 4.8±2.2 & —         & 6.4±0.1 & 4.2±0.4 & 9.6±12.0 & 8.4±9.8 \\
    w-m      & 68.2±18.7 & 91.6±2.8 & \textbf{92.4±2.7} & 84.9±2.3  & 78.3±8.7 & 73.3±17.7 & 87.6±1.3 & 89.3±0.9 \\
    w-m-r    & 15.4±7.8 & 88.7±7.7 & 89.7±5.0 & 83.6±2.1  & 73.8±7.1 & 81.8±2.7 & 86.2±2.4 & \textbf{96.3±2.2} \\
    w-m-e    & 84.9±20.9 & 105.0±7.9 & \textbf{114.4±0.7} & 110.8±0.2 & 109.6±1.0 & 107.9±1.6 & 111.7±1.0 & 112.8±0.8 \\
    w-e      & —         & 114.3±0.6 & 114.7±0.4 & —         & 109.9±1.0 & 108.5±0.5 & 111.7±1.0 & \textbf{117.4±2.8} \\
    \midrule
    \textbf{mean} & — & 82.3±2.1 & 78.9±2.4 & —         & 67.0±3.3 & 65.1±9.3 & 78.5±2.8 & \textbf{83.3±4.4} \\
    \bottomrule
  \end{tabular}
  \label{tb:2}
\end{table*}

To further verify the performance of the proposed method in environments with different dimensional states, we conducted additional experiments in Maze2d with two-dimensional states and sparse rewards, as well as the complex environment Adroit with high-dimensional states. The experimental results are shown in Table \ref{tb:3}. Due to the pseudo-count method based on the multi-codebook VQVAE and the FCM updating, the proposed anti-exploration method achieves the best performance among all the comparative methods in the Maze2d environment and also outperforms the baseline method GPC-SAC in the Adroit environment, which means that the proposed method can obtain stable and good policies in environments with different dimensional states.

\begin{table}[tb!]
  \centering
  \caption{Evaluation on the Maze and Adroit Environments. ``u," ``m," and ``l" denote maze sizes, corresponding to ``umaze," ``medium," and ``large," respectively. The results of PBRL and GPC-SAC are from \cite{shen2024grid}.}
  \resizebox{\linewidth}{!}{ 
  \begin{tabular}{lccccc}
    \toprule
    \textbf{Tasks} & \textbf{CQL} & \textbf{IQL} & \textbf{PBRL} & \textbf{GPC-SAC} & \textbf{OURS} \\
    \midrule
    Maze2d-u   & 56.3  & 46.9  & 86.7  & \textbf{141.0} & 136.9 \\
    Maze2d-m   & 24.8  & 32.0  & 71.1  & 103.7 & \textbf{179.2} \\
    Maze2d-l   & 15.3  & 64.2  & 64.5  & 134.3 & \textbf{231.7} \\
    pen-e      & 107.0 & 117.2 & \textbf{137.7} & 118.8 & 119.9 \\
    hammer-e   & 86.7  & 124.1 & \textbf{127.5} & 95.6  & 85.7 \\
    door-e     & 101.5 & \textbf{105.2} & 95.7  & 101.0 & 103.0 \\
    \midrule
    \textbf{mean} & 65.3  & 81.6  & 97.2  & 115.7 & \textbf{142.7} \\
    \bottomrule
  \end{tabular}
  }
  \label{tb:3}
\end{table}

\subsection{Ablation Studies}
\begin{figure}[!t] 
\centering 
\includegraphics[width=0.45\textwidth]{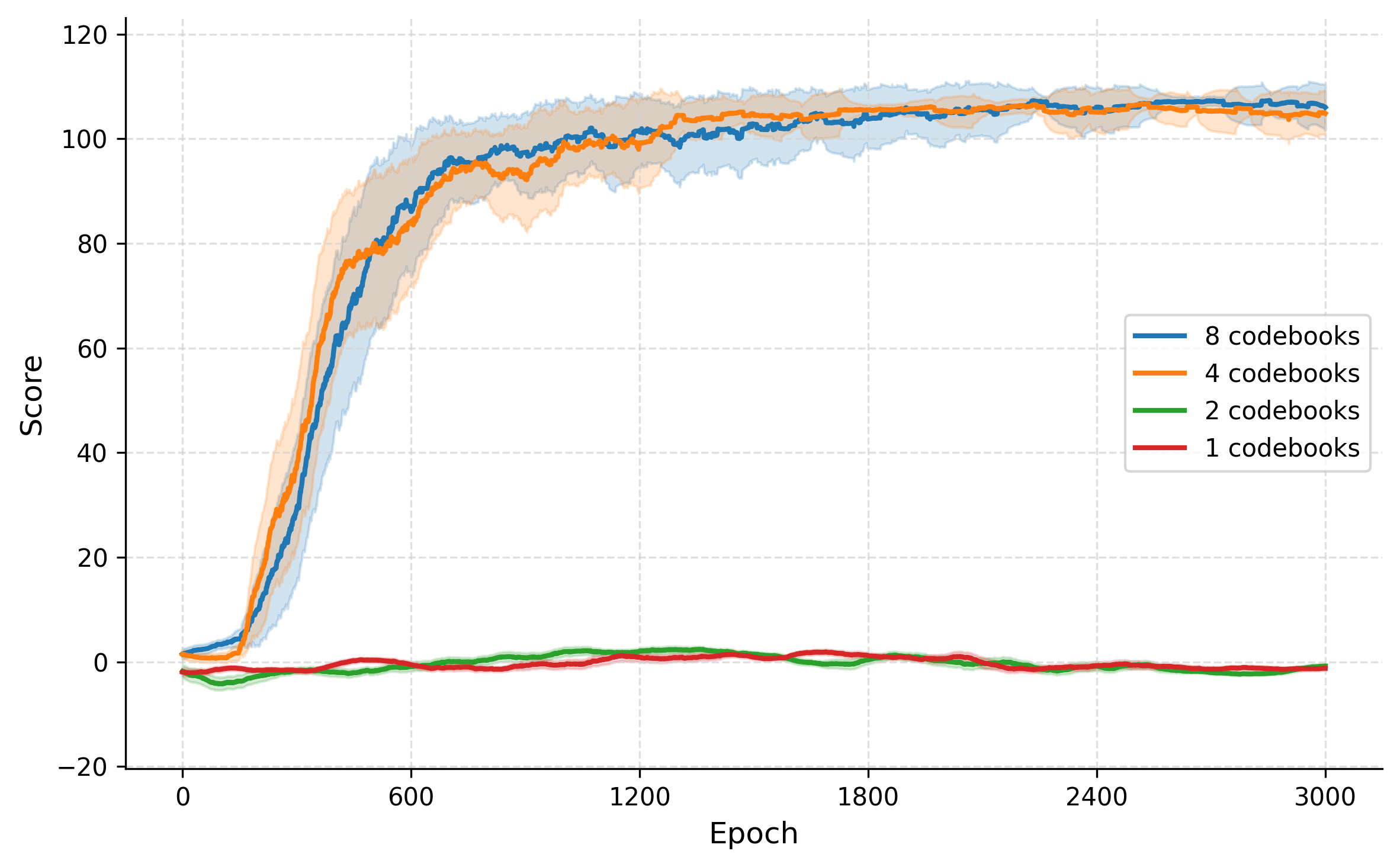} 
\caption{Experimental results with different codebook numbers}
\label{codebook} 
\end{figure}

\textbf{Codebook number:} In this paper, the codebook number $k$ determines the precision of discretization. If $k$ is too small, it will be difficult to accurately represent all the feature information of the input data, while it leads to more parameters and calculation cost if $k$ is too large. In addition, $k$ also needs to be divisible by the dimension of the latent space to ensure a uniform division for the latent space. Therefore, we select $k$ as 8,4,2 and 1 in our experiments, respectively, to verify the influence of the codebook number. Since the halfcheetah task is often used to evaluate the generalization ability on high-reward trajectories and the effect of conservative policy learning \cite{lyu2022mildly}, the expert dataset from this task is used in our ablation experiment. Fig. \ref{codebook} shows the results. The proposed method fails to converge with 1 and 2 codebooks, and the performance with 4 codebooks is similar to that with 8 codebooks, which is because the diversity of the label sequence of vectors used for counting is directly related to the codebook.
It is difficult to effectively extract the features of state-action pairs with too few codebooks, and it will be unable to effectively identify the OOD state-action pairs and provide a correct penalty value. When the codebook number reaches a certain value, further increasing the codebook number will no longer significantly improve the counting accuracy since the vector expression in the codebooks has reached saturation. Instead, it will cause wastes of computing resources. In this case, we design 4 codebooks in our other experiments to balance the accuracy and computing cost. 

\begin{figure}[!t] 
\centering 
\includegraphics[width=0.45\textwidth]{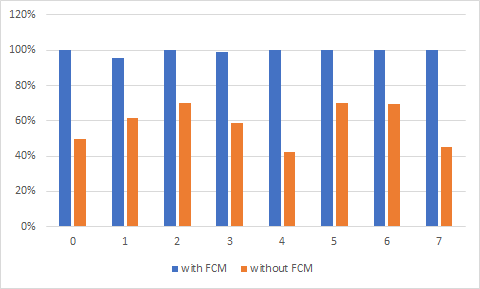} 
\caption{Use rates of codebook vectors with and without FCM.}
\label{usage} 
\end{figure}

\textbf{Fuzzy C-means clustering:} The proposed FCM clustering for codebook update is to improve the use rate of the codebook vectors, handling the information loss issue. To verify the effect of the proposed FCM on improving the use rate of the codebook vectors, the use rates of the codebook vectors are also statistically analyzed in this paper. The proposed VQVAEs with and without FCM are separately trained, and the use rates are calculated when the two well-trained VQVAEs are used to process 100,000 data. 
The experimental results under 8 codebooks are shown in Fig. \ref{usage}, which demonstrates that the use rate of the codebook vectors without FCM is about 60\%, and the use rate is significantly improved, nearly to 100\% when FCM is applied in the multi-codebook VQVAE.

\begin{figure}[!t] 
\centering 
\includegraphics[width=0.45\textwidth]{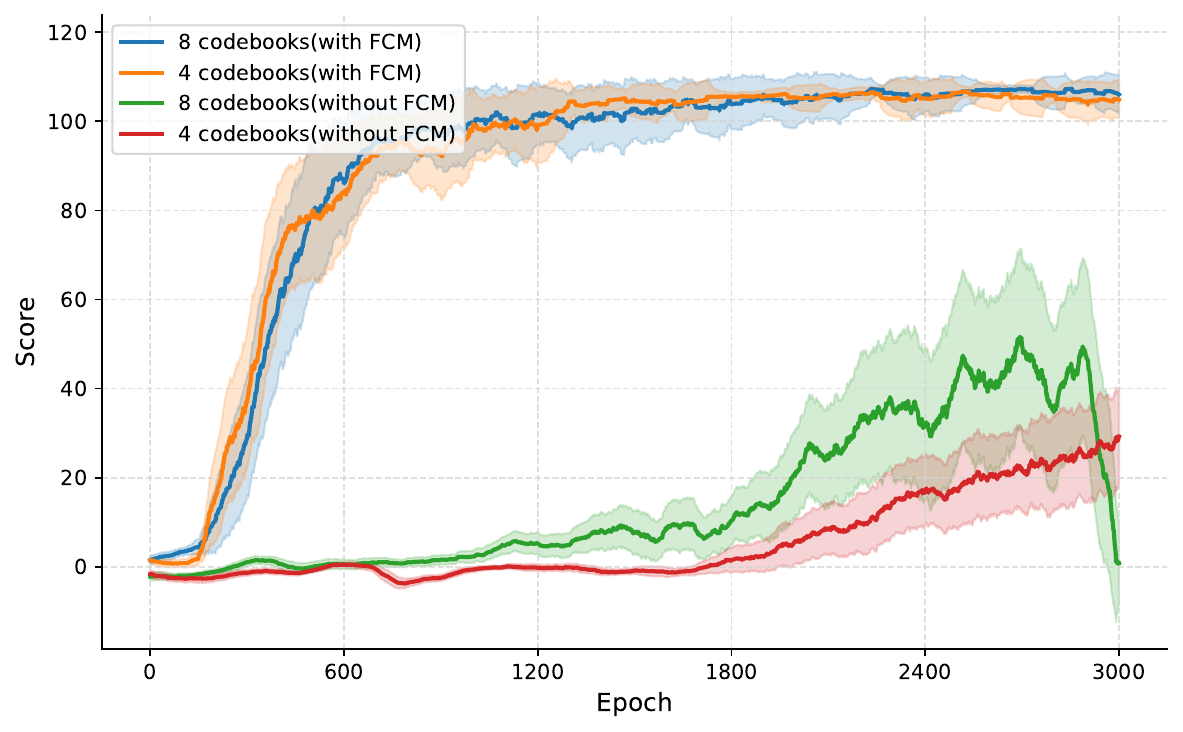} 
\caption{Experimental effects of multi-codebook VQVAE with and without FCM.}
\label{fuzzy} 
\end{figure}

To further verify the effect of FCM on policy training in the proposed anti-exploration method, FCM is removed from the proposed method under the conditions of 8 and 4 codebooks, and the testing experiments are conducted on the halfcheet-expert-v2 dataset. The results are shown in Fig. \ref{fuzzy}. The training processes of the  method without FCM fail to converge under the conditions of both 8 and 4 codebooks due to the low use rate of codebook vectors, resulting in the inability of precisely discretize continuous state-action pairs and information loss. The low-precision pseudo-counting makes it difficult to train the policy. In contrast, the method with FCM can successfully converge and achieve good training results.

\subsection{OOD Data Identification}
\begin{figure}[tb!] 
\centering 
\includegraphics[width=0.45\textwidth]{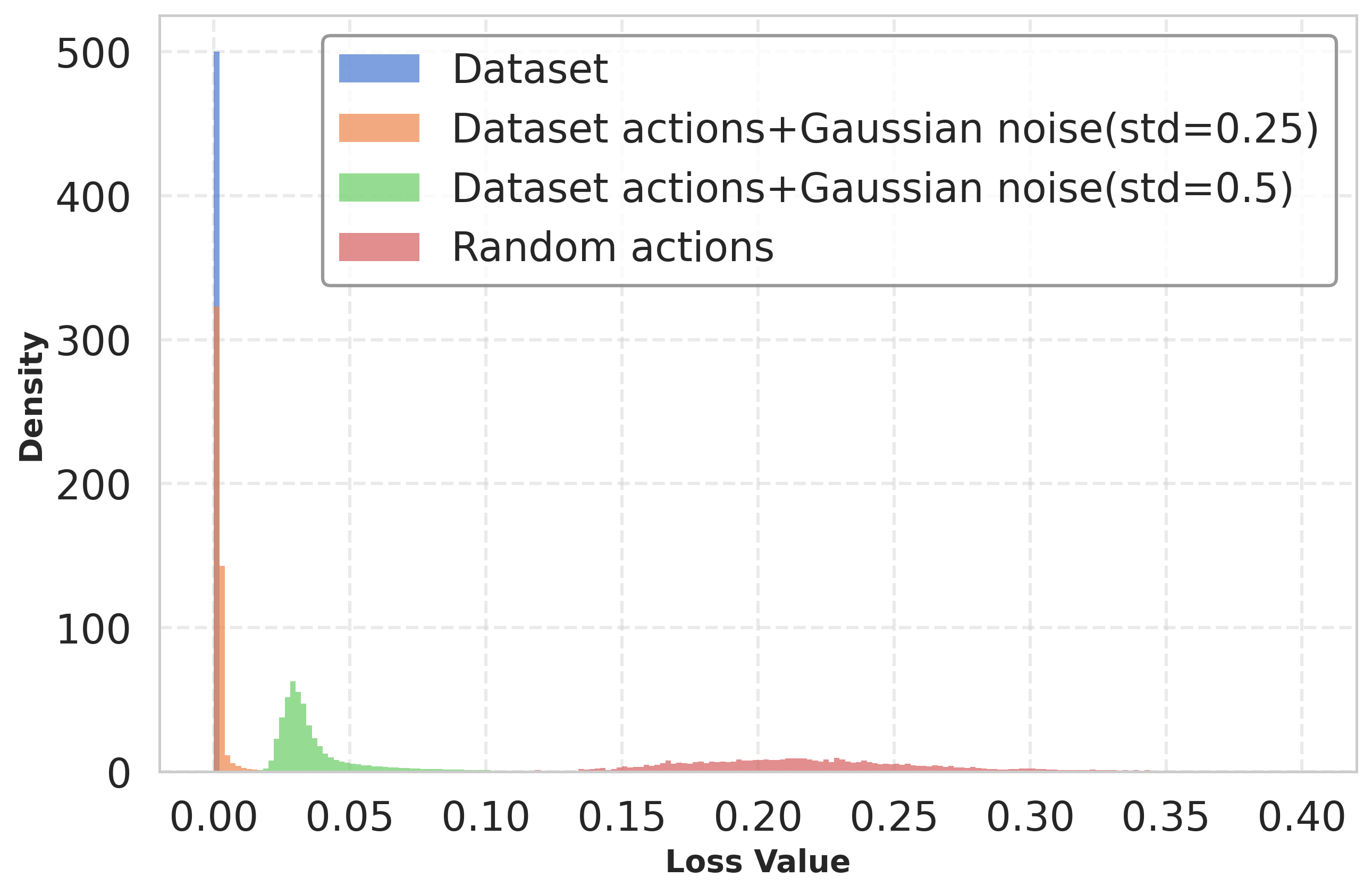} 
\caption{Loss value distribution of VQ-VAE under different OOD data.}
\label{ood1} 
\end{figure}

A purposes of the proposed method is to deal with the OOD issue in offline RL. To verify the ability of the proposed anti-exploration method for identifying OOD state-action pairs, we add two kinds of Gaussian white noises with the same mean 0 but different variances 0.25 and 0.5, respectively, to the offline data, following SAC-RND \cite{nikulin2023anti} and TD3-CVAE \cite{rezaeifar2022offline}. Then, four VQVAE losses of the proposed method, as shown by Eq. \ref{eq:14}, are calculated on the original data without noise, the data with the two kinds of noise, and the randomly generated data, respectively. To visually present the results, we count the numbers of the loss values in different intervals for 100,000 data under the four conditions mentioned above. The histogram results are shown in Fig. \ref{ood1}. When the input data deviate from the original state-action pairs that are also the OOD data, the VQVAE loss value is significantly higher than those of the original data and the low-noise data, which demonstrates the strong identification ability for OOD data. Meanwhile, the loss value for the original data is almost 0, which also demonstrates that the proposed method has a good fitting ability for the offline data.

\subsection{CBF Counting Accuracy Experiment}
\begin{figure}[!t]
\centering

\begin{minipage}{0.48\columnwidth}
\centering
\includegraphics[width=\linewidth]{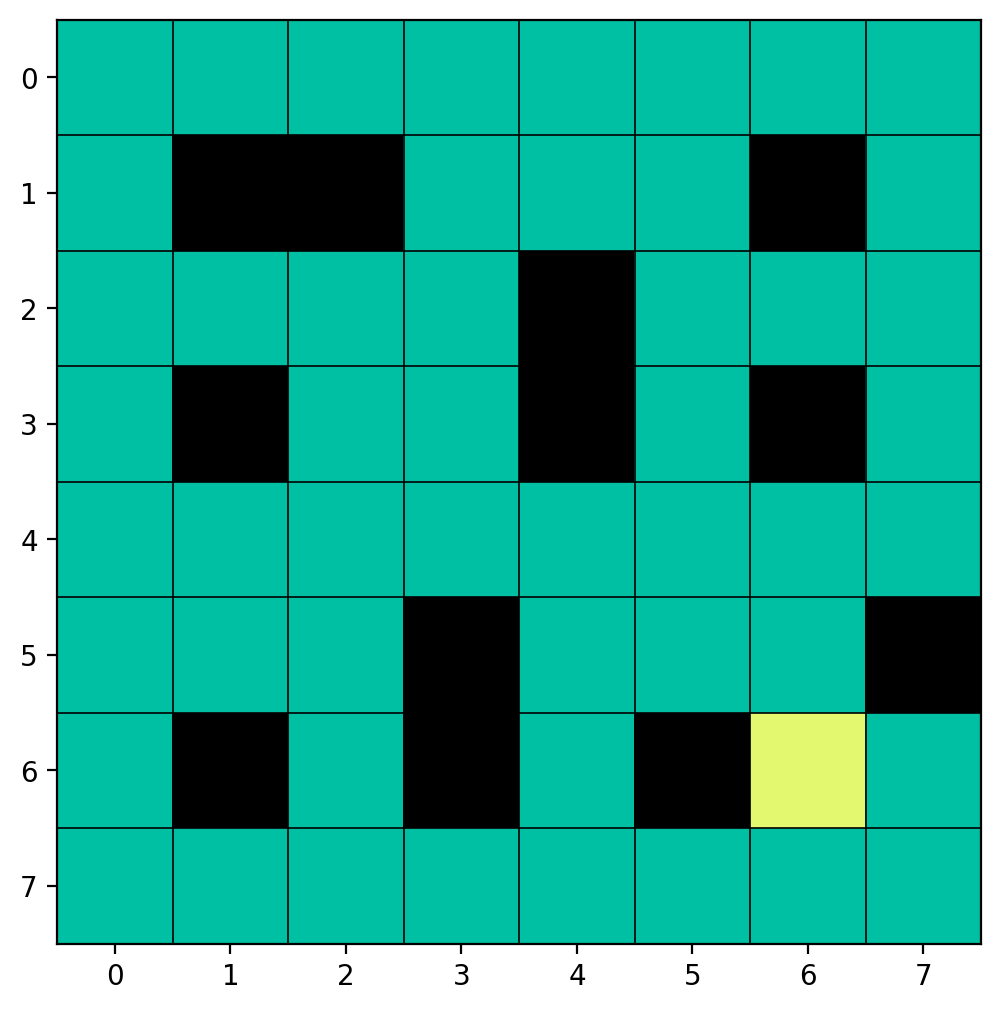}
\vspace{1mm}
(a)
\end{minipage}\hfill
\begin{minipage}{0.48\columnwidth}
\centering
\includegraphics[width=\linewidth]{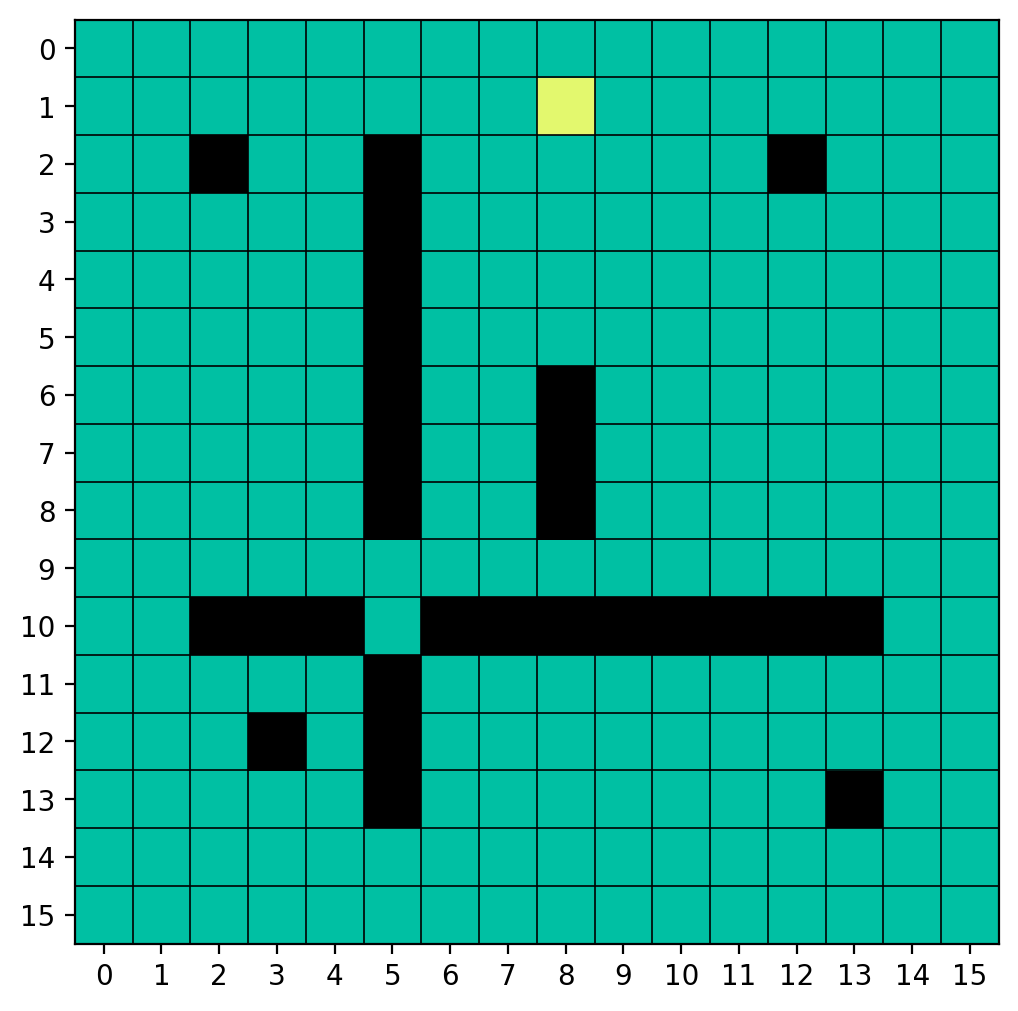}
\vspace{1mm}
(b)
\end{minipage}

\caption{Grid World maps. The green, black, and yellow cells denote the transitable regions, obstacles, and random initial positions of the agent, respectively. (a) $8\times8$. (b) $16\times16$.}
\label{map}
\end{figure}

To verify the effectiveness of the proposed pseudo-count method in terms of counting accuracy, following the setup in Count-MORL \cite{kim2023model}, we evaluated the counting performance of the CBF in our method in the Grid World environment with two grid sizes including $8\times8$ and $16\times16$, as shown in Fig. \ref{map}, and two environment settings including ``with obstacles" and ``without obstacles", resulting in four maps in total. The states in Grid World are the coordinates $(x, y)$ of the grid cells, and the action space consists of four discrete actions: \{up, down, left, right\}. In this case, we can easily obtain the ground truth (GT) of the visitation count of each state–action pair. When the agent executes an out-of-boundary action at the border, its position remains unchanged, but the corresponding state–action pair is still counted. The agent starts from a random position on the map and moves for 10,000 steps under a uniformly random policy, during which the true counts are recorded as the ground truths and the trajectories are saved for the offline dataset, and the offline data are counted by the CBF method in this paper. As shown in Fig. \ref{up}, the heatmap of the estimated visitation counts by CBF is exactly consistent with that of GT, which verifies the effectiveness of CBF as the pseudo-count method in this paper.

\begin{figure}[!t]
\centering

\begin{minipage}{0.31\columnwidth}
\centering
\includegraphics[width=\linewidth]{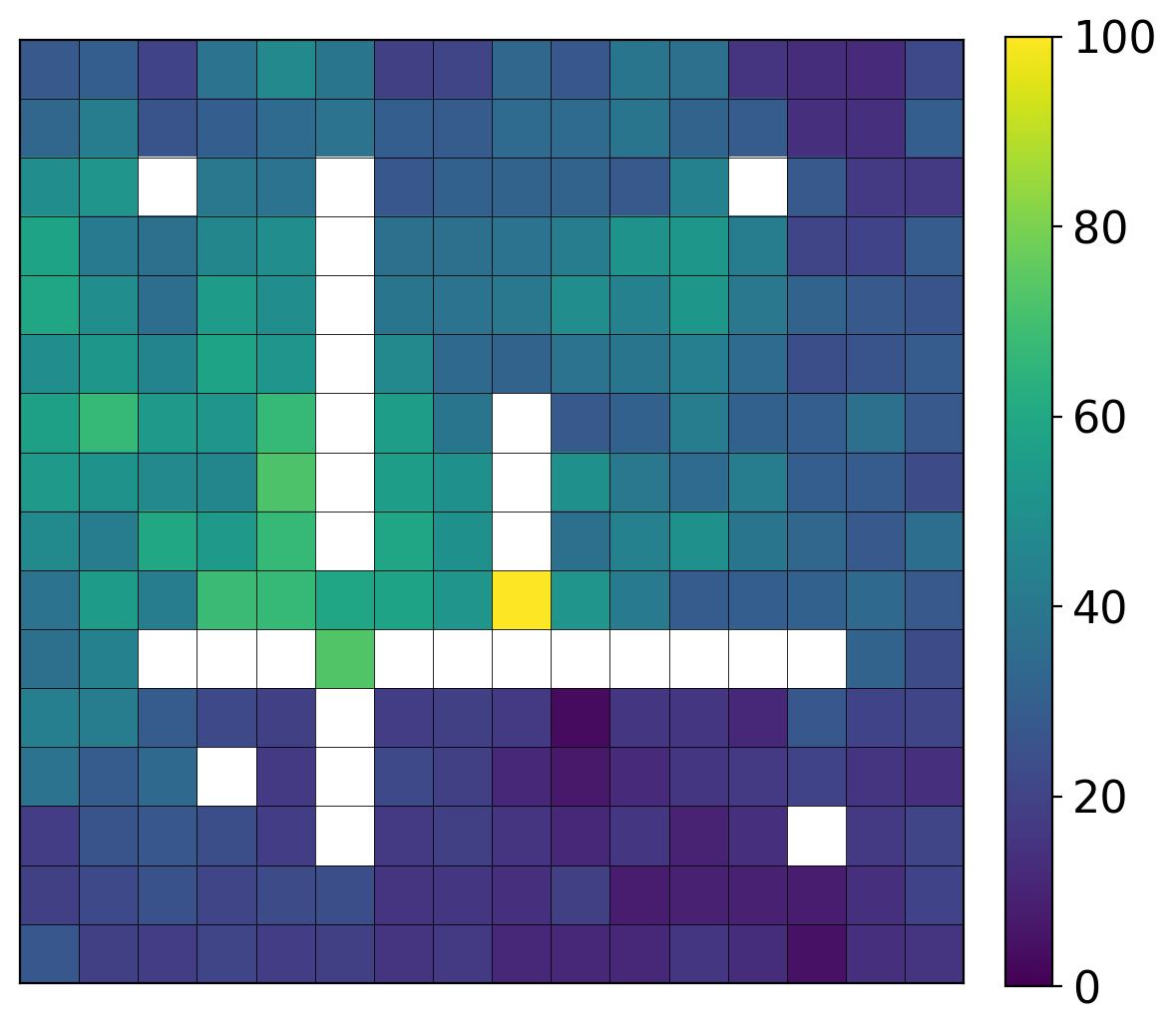}
\vspace{1mm}
(a)
\end{minipage}\hfill
\begin{minipage}{0.31\columnwidth}
\centering
\includegraphics[width=\linewidth]{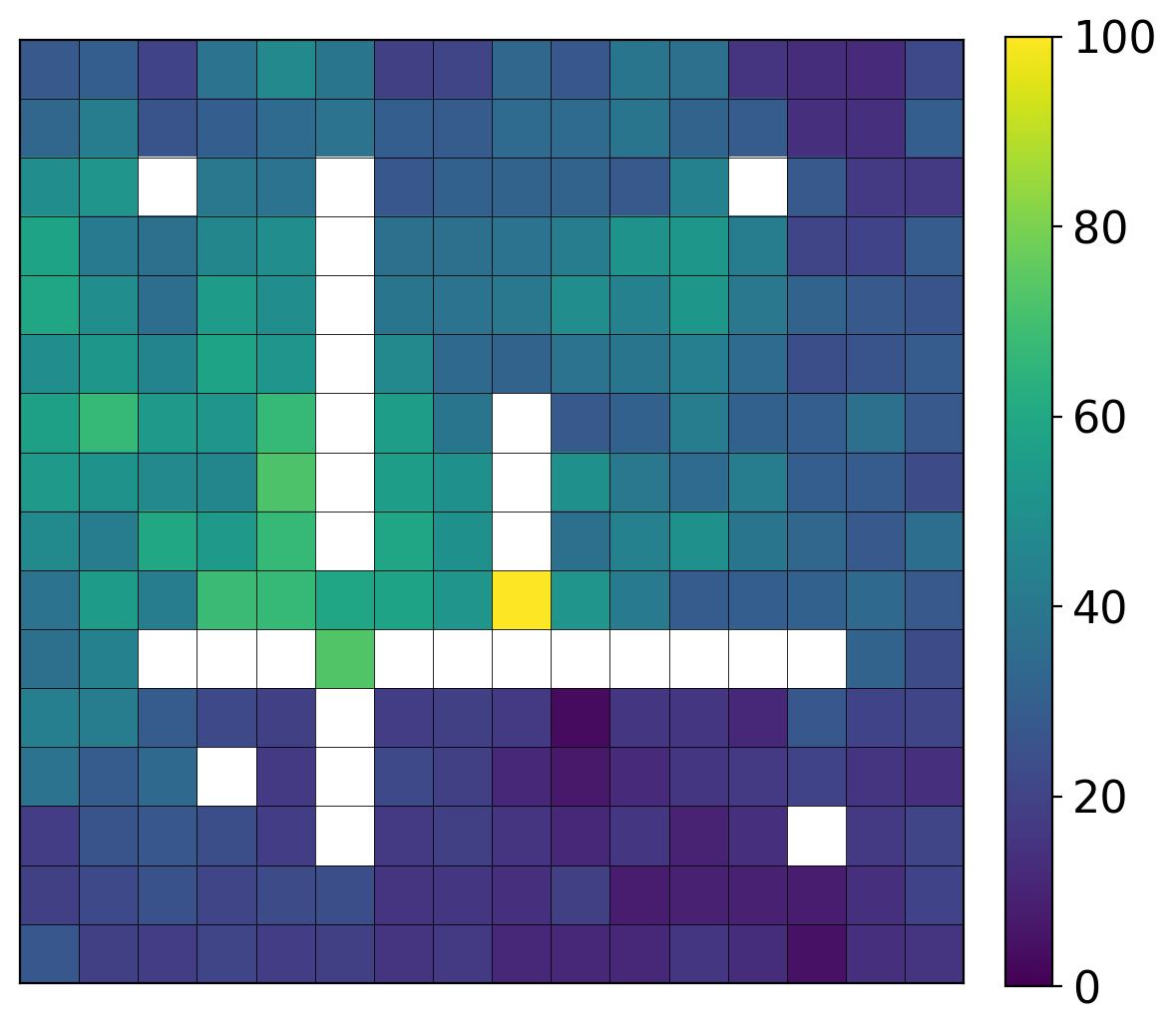}
\vspace{1mm}
(b)
\end{minipage}\hfill
\begin{minipage}{0.31\columnwidth}
\centering
\includegraphics[width=\linewidth]{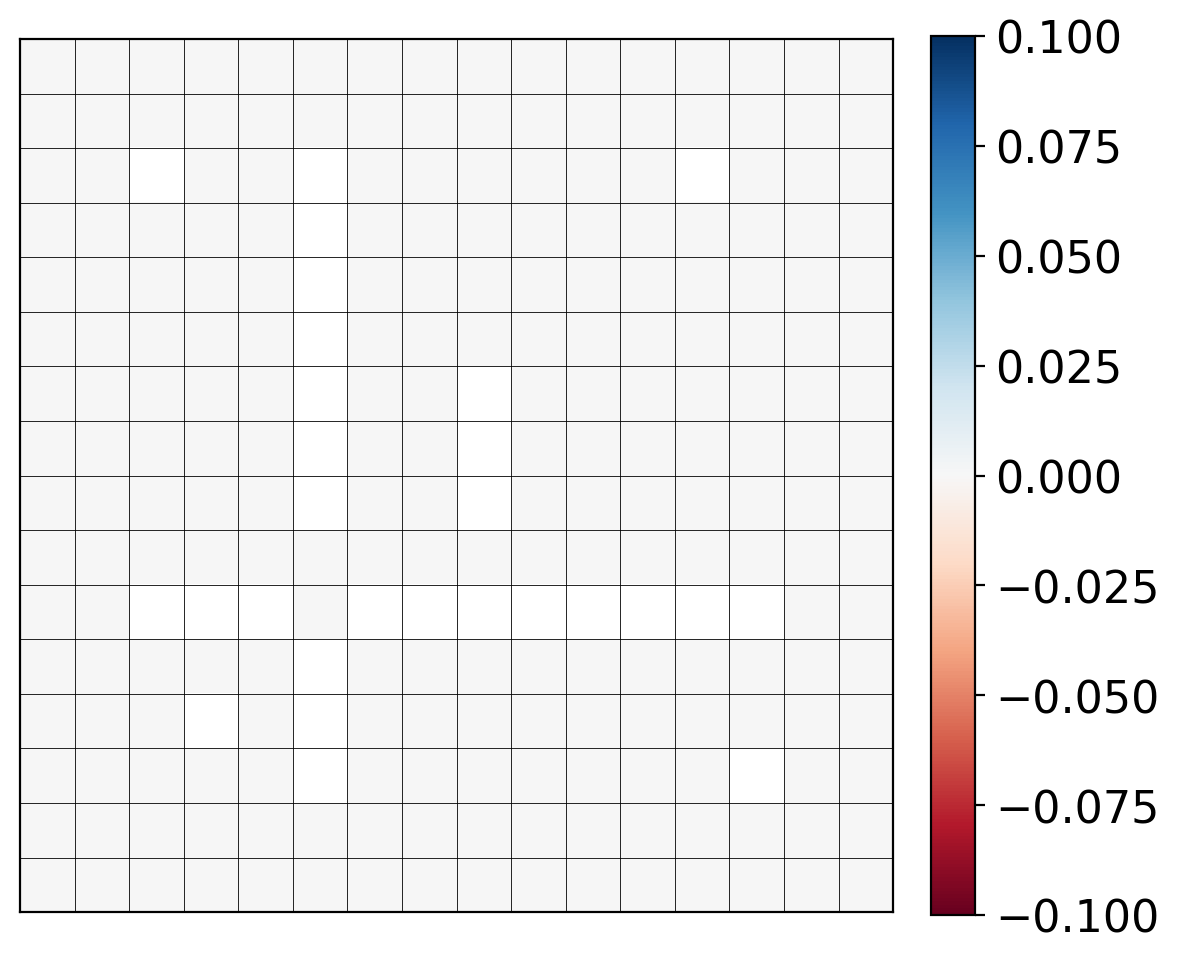}
\vspace{1mm}
(c)
\end{minipage}

\caption{$16\times16$ heatmap comparison for state–action pairs with the action ``up". (a) GT. (b) heatmap of estimated visitation counts. (c) Difference of (a) and (b).}
\label{fig:three_images}
\end{figure}


\subsection{Computing Cost Analysis}
Another purpose of the proposed method is to reduce the computing cost, i.e. to reduce the GPU memory usage in the training stage and runtime in the deployment stage. We calculate the average GPU memory usage as well as the average runtime per step of IQL, CQL, GPC-SAC, SAC, and the proposed method, respectively, under different tasks. The results are shown in Table \ref{tb:4}. Thanks to the efficient counting space utilization in our method which uses the same number of counting containers in all environments, the GPU memory consumption is only 0.2 GB and does not increase with the space dimension of state-action pairs. The runtime of our method is also similar to that of other methods. Compared to the baseline GPC-SAC and other comparative methods, the proposed method achieves the highest scores in multiple D4RL environments while significantly
reducing the computing cost, demonstrating that the proposed method can efficiently handle the dimension disaster issue in offline RL.
 
\begin{table}[tb!]
  \centering
  \caption{Average computing cost of different methods. The reported GPU memory usage represents the average across seven tasks: halfcheetah-medium-expert, walker2d-medium-expert, hopper-medium-expert, maze2d-umaze, pen-expert, hammer-expert, and door-expert. The average runtime per step is measured on halfcheetah-medium-expert over 10,000 environment steps. The implementations of IQL and CQL are based on the official code from \protect\url{https://github.com/tinkoff-ai/CORL}.}
  \renewcommand{\arraystretch}{1.2}
  \setlength{\tabcolsep}{8pt}
  \resizebox{\linewidth}{!}{
  \begin{tabular}{lcc}
    \toprule
    \textbf{Methods} & \textbf{GPU Memory (GB)} & \textbf{Average Runtime (ms)} \\
    \midrule
    IQL       & 0.76 & 0.49 \\
    CQL       & 0.81 & 0.98 \\
    GPC-SAC   & 1.30 &\textbf{0.46} \\
    OURS      & \textbf{0.20}  & 0.56 \\
    \bottomrule
  \end{tabular}
  }
  \label{tb:4}
\end{table}



\section{Conclusions}
To address the issues of dimension disaster and information loss in the process of discretizing continuous state-action pairs in existing pseudo-count methods for offline RL, we propose an anti-exploration method based on the VQVAE and fuzzy C-means clustering. Using VQVAE to discretize continuous state-action pair data, we efficiently reduce the computing cost, and the proposed VQVAE-based method has a good identification ability for OOD data. To improve the use rate of the vectors in codebooks and reduce the information loss in the discretize process, we also develop a pseudo-count framework based on the multi-codebook VQVAE and a vector update method based on FCM, which can not only improve the training efficiency but also the counting accuracy. Comprehensive experiments are conducted on benchmarks consisting of Gym-MuJoCo and Adroit, etc., and experimental results show that the proposed method achieves superior results in terms of policy performance and computing cost compared to SOTA offline RL methods. 
Future work will explore adaptive and multi-scale latent discrete representations, such as dynamically adjusting the VQ-VAE codebook resolution according to offline data density or training stages as well as performing pseudo-counting in latent spaces of different granularities, so as to further balance statistical stability and representational precision, thereby improving robustness against exploration penalties in offline RL.

\bibliographystyle{IEEEtran}
\bibliography{ref}

\vfill

\end{document}